\documentclass[runningheads]{llncs}
\usepackage[T1]{fontenc}
\usepackage{bm}
\usepackage{soul}
\usepackage{hyperref}
\usepackage{graphicx,verbatim}
\usepackage{lineno}
\usepackage{amsmath}
\usepackage{booktabs}
\usepackage{colortbl}
\usepackage{color}
\usepackage{xcolor}
\usepackage{array}
\usepackage{enumitem}
\usepackage[absolute]{textpos}
\begin{document}
\title{\includegraphics[height=1.2em]{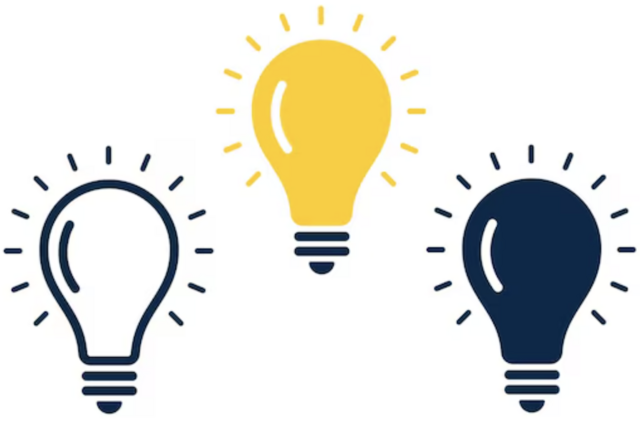} Lamps: Learning Anatomy from Multiple Perspectives via Self-supervision in Chest Radiographs}
%

\author{Ziyu Zhou$^{1,2}$ \quad Haozhe Luo$^{3}$\quad Mohammad Reza Hosseinzadeh Taher$^{2}$ \quad Jiaxuan Pang$^{2}$ \quad Xiaowei Ding$^{\dagger 1}$ \quad Michael B. Gotway$^{4}$ \quad Jianming Liang$^{\dagger 2}$}  
\authorrunning{Ziyu Zhou et al.}
\institute{$^{1}$Shanghai Jiao Tong University \quad  $^{2}$Arizona State University \quad $^{3}$University of Bern \quad $^{4}$Mayo Clinic\\ 
    \email{zhouziyu,dingxiaowei@sjtu.edu.cn \quad zzhou187,mhossei2,jpang12,jianming.liang@asu.edu \quad haozhe.luo@unibe.ch \quad Gotway.Michael@mayo.edu}}

\begin{textblock*}{7cm}(1.8cm,25.5cm)  
    \noindent\rule[0.5ex]{5.2cm}{0.4pt} \\[0.5ex]  
    \footnotesize{$\dagger$ Corresponding author.}
\end{textblock*}

\maketitle              
\begin{abstract}

Foundation models have been successful in natural language processing and computer vision because they are capable of capturing the underlying structures (foundation) of natural languages. However, in medical imaging, the key foundation lies in human anatomy, as these images directly represent the internal structures of the body, reflecting the consistency, coherence, and hierarchy of human anatomy. Yet, existing self-supervised learning (SSL) methods often overlook these perspectives, limiting their ability to effectively learn anatomical features.
    To overcome the limitation, we built Lamps (\textit{\textbf{l}}earning \textit{\textbf{a}}natomy from \textit{\textbf{m}}ultiple \textit{\textbf{p}}erspectives via \textit{\textbf{s}}elf-supervision) pre-trained on large-scale chest radiographs by harmoniously utilizing the consistency, coherence, and hierarchy of human anatomy as the supervision signal. %
Extensive experiments across 10 datasets--evaluated through fine-tuning and emergent property analysis--demonstrate Lamps’ superior robustness, transferability, and clinical potential when compared to 10 baseline models. By learning from multiple perspectives, Lamps presents a unique opportunity for foundation models to develop meaningful, robust representations that are aligned with the structure of human anatomy. All code and pretrained models are publicly released (\href{https://github.com/jlianglab/Lamps}{GitHub.com/JLiangLab/Lamps}).


\end{abstract}
\section{Introduction}

Medical images are highly consistent across scans of the same anatomical region, especially when captured using standardized imaging protocols (e.g., CT, MRI, X-ray). Unlike photographic images, where object positions and shapes can vary significantly, medical images—such as radiological scans—show repeating organs and tissues across patients. As a result, conventional self-supervised learning (SSL) methods developed for photographic images~\cite{caron2021emerging,oquab2023dinov2,grill2020bootstrap} may not be optimal for medical imaging, as they fail to account for the structured and predictable patterns inherent in medical data.
Although several SSL methods have been developed for medical imaging~\cite{Taher_2024_CVPR,zhou2023learning,pang2024asa,zhou2019models}, their performance in interpreting medical images remains limited compared to the success of large language models in understanding natural language~\cite{achiam2023gpt,guo2025deepseek}. We believe this discrepancy arises because SSL methods designed for NLP are adept at capturing the underlying structures (or foundation) of language, allowing intrinsic properties of the language to emerge naturally~\cite{Manning2020}. In contrast, existing SSL methods for medical imaging~\cite{cho2023chess,mh2024lvm} lack the capability to appreciate the foundational structure of medical data—human anatomy. This raises the key question: \textit{How can we effectively learn human anatomy from abundant unlabeled medical images?}

Human anatomy in medical imaging exhibits some unique properties including (1) \textbf{Consistent spatial continuity and coherence}: The relative positions of organs and tissues remain stable due to the anatomical coherence of the human body, ensuring predictable and organized spatial relationships between structures; (2) \textbf{Anatomical structure hierarchy}: Global organs or tissues can be decomposed into smaller anatomical parts (e.g., the left lung is partitioned into superior and inferior lobes). We aim to leverage these properties as intrinsic supervision signals for pretraining in medical imaging, learning anatomical continuity, coherence, and hierarchy from three perspectives.
To learn continuity, we encourage the model to learn the anatomical context exterior to the given structure—for example, given the heart, the model predicts the surrounding lung region. This is termed extrapolation, where the model infers relevant information outside the provided crop without interpolation. For coherence, we guide the model to learn the consistent order of local image patches, reflecting the general consistency in the relative positions of organs and tissues. Finally, we utilize decomposable anatomies to learn hierarchy through composition and decomposition.
In summary, we aim to learn anatomy from the perspectives of continuity, coherence, and hierarchy. To integrate these diverse aspects into a unified foundation model, we employ a teacher-student architecture with an innovative cyclic training strategy that distills knowledge from the student model to the teacher model.

In this research, we focus on chest X-rays (CXRs) due to their cost-effectiveness and high impact in early disease detection~\cite{wang2018prevalence}, and we have collected a large-scale dataset for anatomical learning, as shown in Fig. \ref{fig:mainfigure}-B.
Our contributions are as follows: (1) A novel approach for learning anatomical structures from multiple perspectives using unlabeled CXRs, demonstrating the significant performance enhancement achievable by integrating self-supervised learning (SSL) methods. (2) A new exploration of the emergent properties of pretrained models, including the DNA-test, cross-patient anatomical correspondence, and locality of anatomical structures. (3) A new SSL pretraining method on large-scale data, showing exceptional transferability to various target tasks in medical image analysis.

\section{Learn from Anatomy}

\begin{figure}[t!]
    \centering
    \setlength{\belowcaptionskip}{0.01cm}
    \includegraphics[width=12cm]{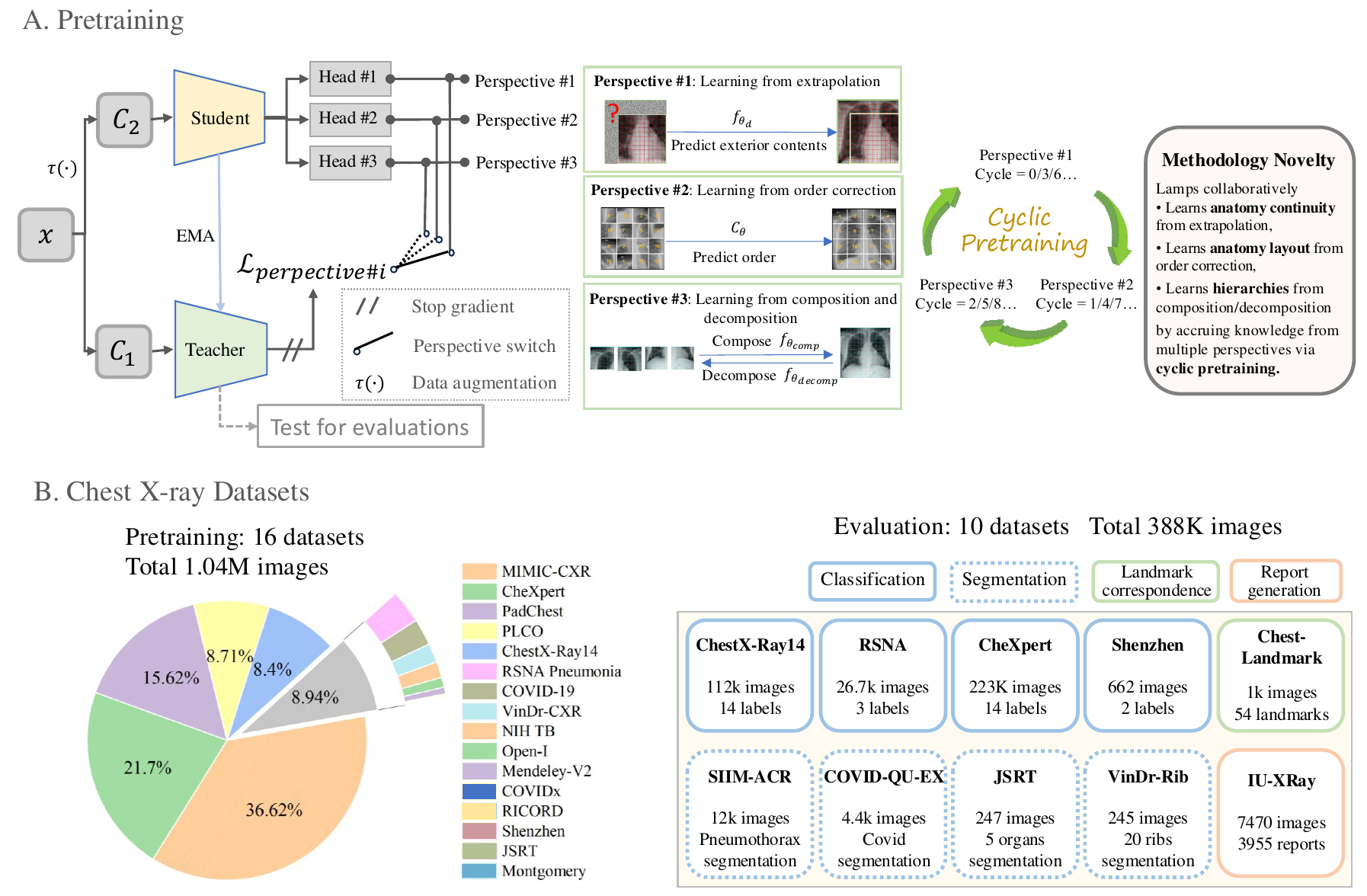}

    \caption{
We introduce Lamps, a foundation model designed to learn from multiple perspectives. A. Lamps is built using our novel framework, which accrues knowledge from three key perspectives: extrapolation, order correction, and composition/decomposition, in a cyclic pretraining process. B. Lamps is pretrained over 1.04M images and comprehensively evaluated on 10 downstream tasks.
    }
    \label{fig:mainfigure}
\end{figure}

Lamps learns from scratch on unlabeled images, with the objective of yielding a common visual representation that is generalizable across disease, organ and modality. As depicted in Fig. \ref{fig:mainfigure}-A, Lamps employs a student-teacher model, trained through a series of learning perspectives with \textit{cyclic pretraining} to accrue anatomical knowledge. Once trained, the teacher model can be used independently for fine-tuning or zero-shot evaluations. In the following, we introduce the three learning perspectives and integrate them into one training framework.

\noindent \textbf{Perspective \#1: \ul{learning anatomical continuity via extrapolation.}} In this context, extrapolation refers to the model using information from a given image crop to predict semantically relevant details \textit{outside} of that crop based on patterns learned from similar images. Our method differs significantly from interpolation and existing MIM approaches~\cite{he2022masked,xie2022simmim} in two key ways: (1) it does not rely on anatomical cues from unmasked patches surrounding the masked areas, resulting in more robust representations; and (2) it restores the features of the masked region rather than the original image pixels, encouraging the encoder to learn high-level representations for anatomical understanding rather than low-level superficial features for image restoration. 
    Image pre-processing details:
An input image $x$ is divided into a grid $G$ of $18\times18$ non-overlapping patches. We extract a random crop $C_1$ with $n_1\times n_1$ patches from $G$, and an random crop $C_2$ with $n_2\times n_2$ patches inside $C_1$, where $n_2 < n_1$, and $C_1$ and $C_2$ always begin from one of the nodes in $G$. $C_1$ is regarded as the ground truth and converted into a sequence of $N_1$ non-overlapping patches, where $N_1=n_1^2$, then fed to \texttt{Teacher} to obtain corresponding patch-level embeddings $\bm{e}_t = {\bm{e}_{t_1}, ..., \bm{e}_{t_{N_1}}}$ where $\bm{e}_{t_k}$ is the embedding associated with the $k$-th patch. Similarly, $C_2$ is input to \texttt{Student} and $\texttt{Head\#1}$ $f_{\theta_d}$to get the predicted embeddings $\bm{e}_s= f_{\theta_d}(f_{\theta_s}(C_2, M))$, where  $\bm{e}_s = {\bm{e}_{s_1}, ..., \bm{e}_{s_{N_1}}}, M \in \{0,1\}_{n_1 \times n_1}$ denotes the set of masked patches. An $l_1$ loss is employed on the masked patches to align the predicted embeddings from \texttt{Student} with the ground truth embeddings from \texttt{Teacher}:
\begin{equation}\label{eq:extrapolation}
    \mathcal{L}_{extrap} = \left\| \bm{e}_s[M]-\bm{e}_t[M] \right\| _1
\end{equation}

\noindent \textbf{Perspective \#2: \ul{learning anatomical coherence via order correction.}} In medical imaging (e.g., chest X-ray), the relative positions of organs and tissues are consistent due to anatomical coherence. As such, the relative patch order is a strong signal for patch-level supervision. For implementation, $C_1$, as extracted in the last section, is randomly shuffled and input to \texttt{Student}, while the original $C_1$ (patches in order) is input to \texttt{Teacher} to get corresponding patch-level embeddings $\bm{e}_s$ and $\bm{e}_t$, respectively. $\bm{e}_s$ is then passed to a prediction classifier $\texttt{Head\#2}$ $C_{\theta}$ to get the predicted order $P^o$ of the shuffled patches. Besides, the shuffled patched embedding from \texttt{Student} should be consistent with the embeddings from \texttt{Teacher}. The total loss of this branch will be order correction loss and local consistency loss as in Eq.~\ref{eq:shuffle}, where $T^o$ is the target order of the shuffled patches, $CE(a,b)=-a \log b$ is used to compute the order classification loss. $\bm{e}_s$ represents the student's patch embeddings in shuffled order, while $\bm{e}_t$ is the teacher's embeddings in the original order. We reorder $\bm{e}_t$ as $\bm{e}_t[T^o]$ and use MSE loss to align the shuffled patch embeddings from \texttt{Student} to {\texttt{Teacher}'s}.
\begin{equation}\label{eq:shuffle}
    \mathcal{L}_{shuffle} = \lambda * CE(T^o, P^o)+MSE(\bm{e}_s, \bm{e}_t[T^o])
\end{equation}

\noindent \textbf{Perspective \#3: \ul{learning anatomical hierarchy via composition and decomposition.}} From a hierarchical view in anatomies, organs and tissues can be broken down into smaller components. This decomposition requires the ability to distinguish anatomical and pathological structures, allowing the model to learn the relationships between the whole anatomy and its smaller components. In detail, to learn the anatomical structure in a hierarchical way, the initial image is randomly cropped to $C_1$ with a cropping ratio $r=exp(-\frac{epoch}{t})$, where $t$ increases as training progresses, enabling a coarse-to-fine crop. Inspired by~\cite{Taher_2024_CVPR}, $C_1$ is divided into 4 sub-crops $C_2 = \{C_1^i\}_{i=1}^4$. To capture composition, the sub-crops are encoded by \texttt{Student} to get corresponding embeddings $\{\bm{e}_{s_i}\}_{i=1}^4$, which are concatenated and input to a linear layer $\texttt{Head\#3}$ $f_{\theta_{comp}}$ to aggregate the features: $\bm{e}_{comp}=f_{\theta_{comp}}(\oplus(\{\bm{e}_{s_i}\}_{i=1}^4))$, where $\oplus$ is the concatenation operator. The composition loss will be: $\mathcal{L}_{comp} = \left\| \bm{e}_{comp}-\bm{e}_t \right\| _1$, where $\bm{e}_t$ is the embedding of $C_1$ encoded by \texttt{Teacher}. Symmetrically, to learn decomposition, $C_1$ is encoded by \texttt{Student} to get $\bm{e}_s$, and then input to $f_{\theta_{decomp}}$ to decompose into 4 sub-features: $\bm{e}_{decomp}=f_{\theta_{decomp}}(\bm{e}_s)=\{\bm{e}_{decomp_i}\}_{i=1}^4$. the decomposition loss will be $\mathcal{L}_{decomp} = \frac{1}{4}\sum_{i=1}^4 \left\| \bm{e}_{{decomp}_i}-\bm{e}_{t_i} \right\| _1$, where $\{\bm{e}_{t_i}\}_{i=1}^4$ are the embeddings of $\{C_1^i\}_{i=1}^4$ encoded by \texttt{Teacher}. The total loss in this branch will be:
\begin{equation}\label{eq:compdecomp
}
    \mathcal{L}_{comp-decomp} = \mathcal{L}_{comp}+\mathcal{L}_{decomp}
\end{equation}

\noindent \textbf{\ul{Training pipeline.}} Training a model with a single optimization objective can only capture a limited understanding. Directly combine multiple objectives, where the total loss in each iteration is the sum of the individual losses, the model will become conflated about its optimization targets as different loss functions lead to divergent optimization directions. Therefore, we integrate the 3 perspectives into a unified framework by utilizing \textit{cyclic pretraining}~\cite{ma2023foundation}, which reduces computational costs compared to co-training within a single epoch and avoids the complexity of generating heterogeneous data for different tasks. For the cyclic pattern, each learning perspective will be trained once in every 3 epochs: if $epoch\%3==0$, $\mathcal{L}=\mathcal{L}_{extrap}$; else if $epoch\%3==1$, $\mathcal{L} =\mathcal{L}_{shuffle}$; else $epoch\%3==2$, $\mathcal{L} =\mathcal{L}_{comp-decomp}$.


\section{Experiments and Results}

\noindent \textbf{\S3.1 \ul{Experiment protocol.}} Our Lamps takes the base version of the Swin transformer (Swin-B)~\cite{liu2021swin} as the backbone. We adopt~\cite{caron2021emerging} for optimization settings (e.g., learning rate schedule, optimizer, etc.) and teacher weight updates. In Lamps, each learning branch utilizes the same student-teacher encoder architecture, with unique heads tailored to distinct learning perspectives (see Fig. \ref{fig:mainfigure}-A). 
For perspective \#1, $n_1=14, n_2=11$, $\texttt{Head} \#1$ ($f_{\theta_d}$ is an 8-layer attention block) 
reconstructs the embeddings of masked patches. 
In perspective \#2, $\lambda=0.1$, $\texttt{Head} \#2$ ($C_{\theta}$ is a linear layer). 
In the last perspective \#3, $\texttt{Head} \#3$ ($f_{\theta_{comp}}$ and $f_{\theta_{decomp}}$ are three-layer MLP heads). The input image size is $448^2$. To demonstrate the scalability of our framework, we pretrain Lamps on both ChestX-ray14~\cite{wang2017chestx} (86K images) and a large corpus of 1.04M images for 100 epochs, shown in Fig. \ref{fig:mainfigure}-B.
We compare Lamps against 10 representative publicly available methods: vision foundation models Ark~\cite{ma2023foundation}, RAD-DINO~\cite{perez2024rad}, Adam-v2~\cite{Taher_2024_CVPR} pretrained on 704K, 838K and 1M chest radiographs, vision-language models KAD~\cite{zhang2023knowledge}, DeViDe~\cite{luo2024devide} pretrained on MIMIC~\cite{johnson2019mimic} dataset, and vision SSL models DINO~\cite{caron2021emerging}, BYOL~\cite{grill2020bootstrap}, SelfPatch~\cite{yun2022patch}, POPAR~\cite{pang2022popar} and PEAC~\cite{zhou2023learning} pretrained on ChestX-ray14~\cite{wang2017chestx}. We evaluate Lamps across various settings for segmentation, classification and report generation as shown in Fig. \ref{fig:mainfigure}-B. Specifically, we investigate the \textit{emergent} properties (localizability, anatomy correspondence and DNA-test) of Lamps using Chest-Landmark dataset (1000 test images from ChestX-ray14 dataset, manually annotated for different anatomical landmarks).



\smallskip
\noindent \textbf{\S3.2 \ul{Lamps exhibits emergent properties going beyond current SSL methods.}} The emergent properties of foundation models are some capabilities are not explicitly programmed or anticipated during initial training. We highlight our framework’s anatomy understanding capabilities by examining the unique emergent properties of Lamps’ embeddings across various \underline{zero-shot} settings.

\begin{figure}[thp]
    \centering
    \setlength{\belowcaptionskip}{0.01cm}
    \includegraphics[width=11.5cm]{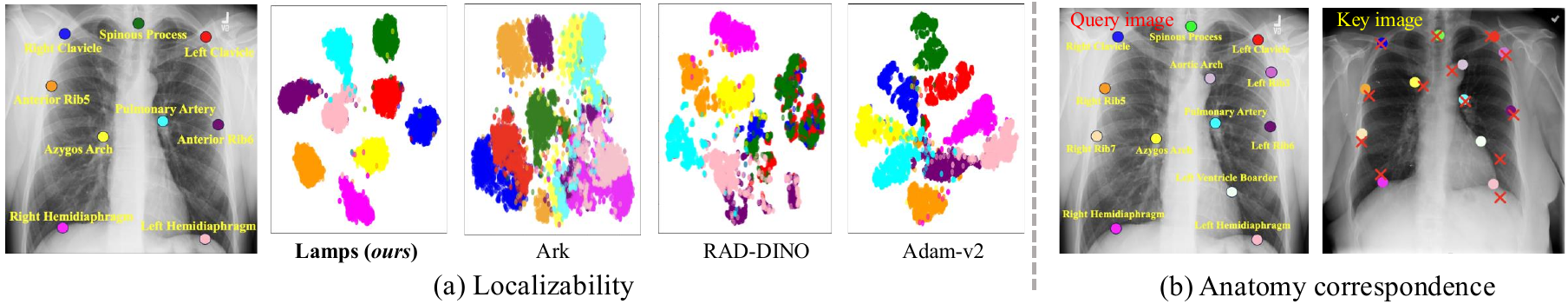}

    \caption{
Lamps demonstrates anatomical understanding through emergent properties: (a) Localizability: distinguishing anatomical structures across patients; (b) Anatomy correspondence: accurately matching identical anatomies across patients (zero-shot predictions (red crosses) vs. ground truth (colored circles)).
    }
    \label{fig:locality_correspondence}
\end{figure}

\smallskip
\noindent \textbf{\textit{(1) Localizability.}} Chest radiographs display consistent anatomical structures due to standardized imaging protocols and stable anatomy across individuals. We explore the foundation model’s ability to convert each pixel into semantically rich embeddings, where different anatomical structures are linked to distinct embeddings, and the same structures have nearly identical embeddings across patients. To assess this, we compile the Chest-Landmark dataset with 9 anatomical landmarks (see Fig. \ref{fig:locality_correspondence}-a). To assess this, we compile the Chest-Landmark dataset with 9 anatomical landmarks (see Fig. \ref{fig:locality_correspondence}-a). We extract local embeddings at the landmark location, corresponding to a $32^2$ resolution, from $448^2$ crops centered around each landmark in the $1024^2$ original images. These embeddings are visualized using a t-SNE~\cite{van2008visualizing} plot. In Fig. \ref{fig:locality_correspondence}-a, Lamps effectively distinguishes anatomical landmarks, while other models such as Ark, RAD-DINO, and Adam-v2 produce confusing embeddings, demonstrating Lamps’ ability to consistently represent semantically similar structures across patients.

\smallskip
\noindent \textbf{\textit{(2) Anatomy correspondence.}} Identifying the same anatomical structures consistently across patients helps ensure that diagnoses are based on reliable, reproducible features. To demonstrate the efficacy of our Lamps in capturing a diverse range of anatomical structures, we utilize patch-level features to query the same anatomy across different patients in zero-shot setting. In detail, we use Chest-Landmark dataset and choose $N_q=13$ landmarks as shown in Fig. \ref{fig:locality_correspondence}-b. For a given query image, we get the chosen landmarks' features $E_q = \{E_q^i\}_{i=1}^{N_q}$ the same with \textit{Sec. (1) Localizability}. Then for the paired key image, we extract $N_k$ patches by sliding a window of size $448^2$ with a stride of 8 (zero padding for the boundary patches), then input these patches to the pretrained backbones to get a dictionary of features for the key image $E_k = \{E_k^j\}_{j=1}^{N_k}$. Finally, for each query landmark feature in $E_q$, we find the most similar feature in $E_k$ with $l_2$ distance and the position in the key image is the predicted corresponding landmark. We compute the errors between the 13 predicted landmarks and the ground truth across the studied baselines. The errors are averaged for the 13 landmarks as shown in Tab. \ref{tab:anatomy matching} and a pair of image landmark matching is visualized in Fig. \ref{fig:locality_correspondence}-b. The results demonstrate that Lamps-encoded features accurately detect anatomical landmarks, capturing specific regions and maintaining consistency despite significant morphological variations.

\begin{table}[h!]

        \centering
        \caption{Anatomical structure matching error across different chest X-rays pretrained models on size of $1024^2$ images.}
        \label{tab:anatomy matching}
        \resizebox{0.95\columnwidth}{!}{
        \begin{tabular}{c|c|c|c|c|c|c}
        \toprule    
        \rowcolor{gray!20} Lamps& RAD-DINO &Adam-v2& KAD & DeViDe  &POPAR& PEAC\\     
        $\mathbf{68.17\pm 44.45}$ & $120.28\pm 150.31$ &$\underline{94.03\pm 131.18}$& $296.52\pm 263.65$& $188.21\pm 226.31$ & $120.69\pm 134.32$ & $95.49\pm 114.22$\\
        
        \bottomrule
        \end{tabular}}
    
\end{table}

\begin{figure}[htb]
    \centering
    \begin{minipage}[t]{0.53\linewidth}
        \vspace{0pt} 
        \centering
        \includegraphics[width=\linewidth]{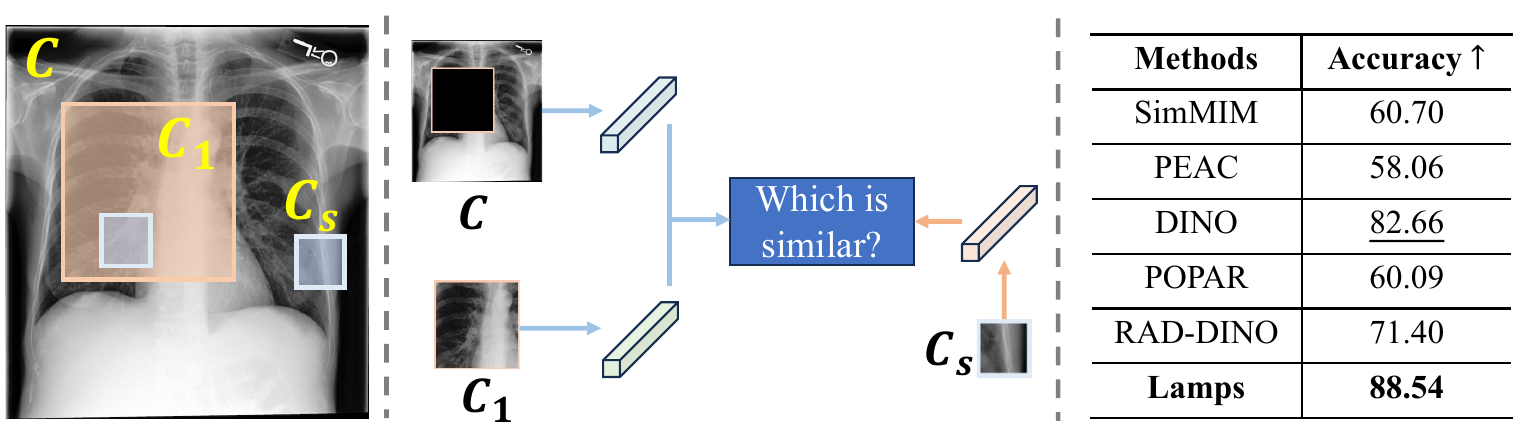}
        \caption{Lamps learns hierarchical anatomies. Our model leverages DNA-test data to encode part-whole structures, enabling discrimination of whether a part structure belongs to its corresponding whole.}
        \label{fig:dnatest}
    \end{minipage}%
    \hfill 
    \begin{minipage}[t]{0.4\linewidth}
        \vspace{0pt} 
        \centering
        \includegraphics[width=\linewidth]{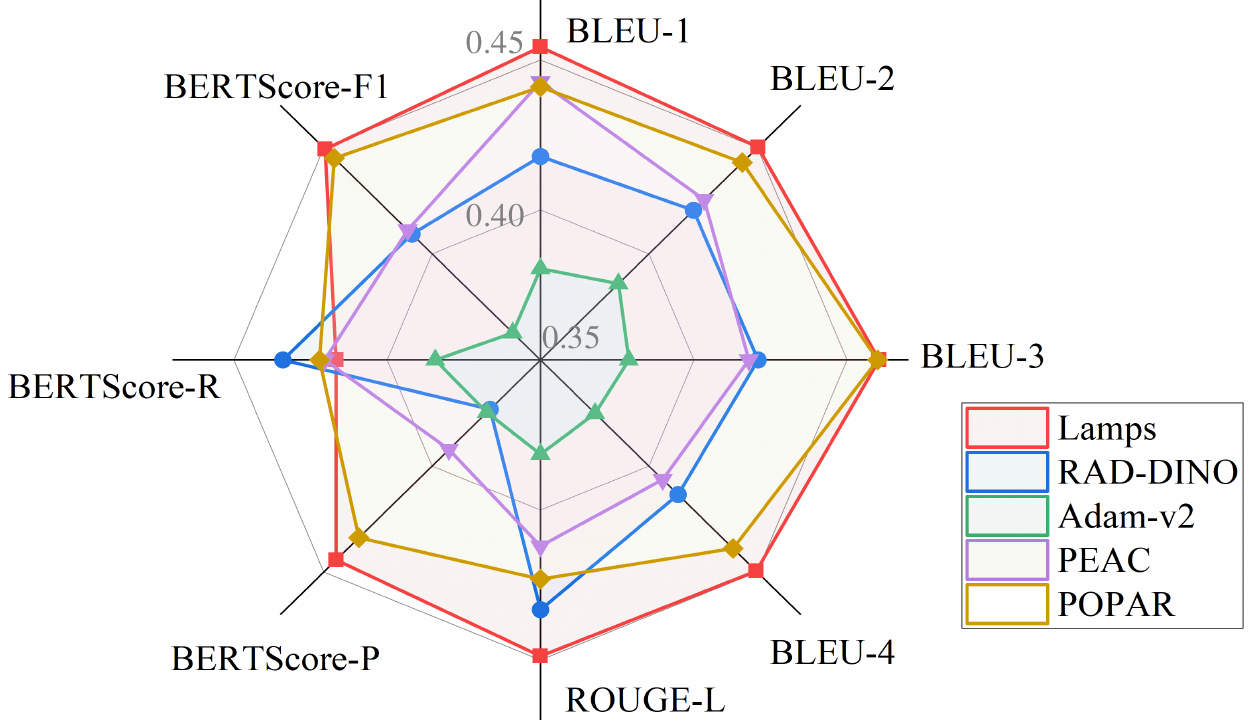}
        \caption{Lamps outperforms baselines on 7 of 8 metrics for report generation.}
        \label{fig:report_generation}
    \end{minipage}
\end{figure}

\noindent \textbf{\textit{(3) DNA-test.}} We define the \textit{DNA-test} to assess a model’s ability to recognize part-whole relationships in anatomical structures, a key aspect of hierarchical understanding. For instance, it evaluates whether the model can correctly associate smaller anatomical parts (e.g., a lung lobe) with their larger structures (e.g., the entire lung). A model that can accurately discriminate part-whole relationships is likely to have a deeper and more nuanced understanding of anatomical structures. As illustrated in Fig. \ref{fig:dnatest}, we crop a global structure $C_1$ from the original image and define $C$ as the remaining area. A smaller region, $C_s$, is randomly cropped from either $C_1$ or $C$. These regions are then encoded by the pretrained model (without fine-tuning) to obtain embeddings $E_C$, $E_{C_1}$, and $E_{C_s}$. We compute cosine similarity between $E_{C_s}$ and both $E_C$ and $E_{C_1}$, selecting the closest match to evaluate accuracy on the ChestX-ray14 test set. As shown in Fig. \ref{fig:dnatest}, Lamps achieves the highest accuracy, demonstrating its superior ability to capture hierarchical anatomical relationships which ensures better interpretability of the learned high-level features.

\begin{figure}[h!]
\begin{minipage}[c]{1\textwidth}
    \centering
    \includegraphics[width=12cm]{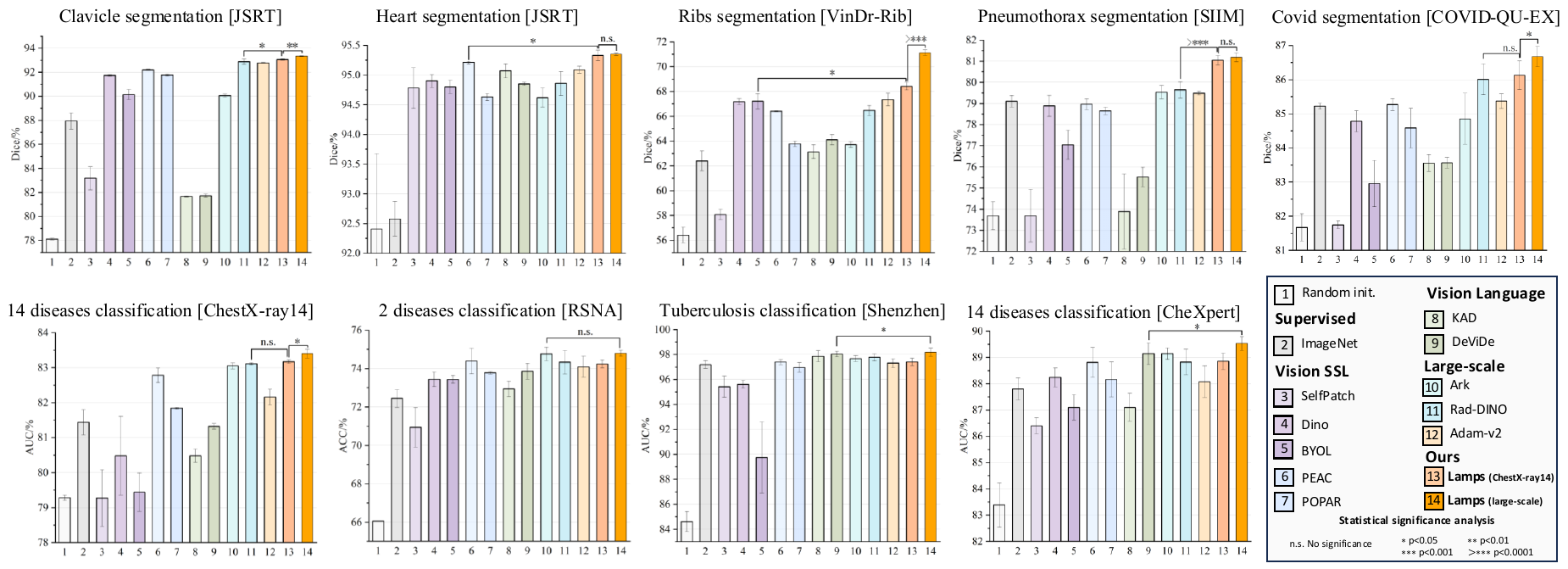}

    \caption{
Lamps achieves superior generalization and robustness, surpassing SOTA SSL methods on diverse downstream tasks. For each task, statistical significance testing ($p$ < 0.05) was performed to compare Lamps with the best-performing SSL baseline.
    }
    \label{fig:finetune}
\end{minipage}
\end{figure}

\smallskip
\noindent \textbf{\S 3.3 \ul{Lamps excels in full transfer learning.}} We investigate the generalizability of Lamps's representations in full fine-tuning settings across a wide range of downstream tasks. To this end, in addition to training from scratch (the lower bound baseline) and a fully supervised ImageNet model, we compare two versions of Lamps---one pretrained on 86K images from the ChestX-ray14 dataset and the other on 1M chest X-ray images from 16 publicly available datasets---against 10 SOTA self-supervised and fully supervised baselines, spanning 3 organ segmentation tasks (clavicle, heart, and rib), 2 disease segmentation tasks (Pneumothorax and COVID), and 4 disease classification tasks (common thoracic diseases). As shown in Fig. \ref{fig:finetune}, across all organ and disease segmentation tasks, Lamps consistently excels beyond all fully supervised and self-supervised baselines. Similarly, for classification tasks, Lamps demonstrates competitive performance in one task and significantly outperforms the baselines in the remaining tasks. In our evaluation for radiology report generation on the IU-XRay dataset~\cite{demner2016preparing}, as shown in Fig. \ref{fig:report_generation}, Lamps outperforms the SSL baselines, achieving the highest scores in 7 out of 8 metrics. In summary, the consistent performance gains over baselines across various classification, segmentation, and report generation tasks underscore the versatility and generalizability of Lamps' learned representations.



\begin{table}[h]
    \vspace{-0.6cm}
        \centering
        \caption{Ablations of individual learning perspectives and combining manners. }
        \label{ablation:components}
        \resizebox{1.0\textwidth}{!}{
        \begin{tabular}{c|c|cc|cc}
        \toprule    

         Learning& Combination  & \multicolumn{2}{c|}{Finetuning} &  \multicolumn{2}{c}{Anatomical understanding}\\
        \cline{3-6}
         perspective & manner & ChestX-ray14 & SIIM & DNA-test & Corr-error \\
        \midrule
        Extrapolation&- & $82.44 \pm 0.26$ & $80.44 \pm 0.32$ & 78.70 & 69.33\\
        Order correction&- & $82.06 \pm 0.26$ & $79.64 \pm 0.18$ & 69.42 & \textbf{65.65}\\
        Comp-decomp&- &$81.55 \pm 0.21$ & $79.73 \pm 0.76$ & \underline{86.51} & 197.27\\
        All&Directly & $\underline{82.96 \pm 0.24}$& $\underline{80.77 \pm 0.10}$ & 60.29 & 71.80\\
        All&Cyclic training & $\mathbf{83.16 \pm 0.05}$& $\mathbf{81.05 \pm 0.22}$ & \textbf{88.54} & \underline{68.17}\\

        \bottomrule
        \end{tabular}}
    \end{table}

\smallskip
\noindent \textbf{\S 3.4 \ul{Ablation: effectiveness of the learning objectives.}} We evaluate the impact of each learning perspective in Lamps by testing individual learning perspectives and different combination methods. We fine-tune the models on the 14-class thoracic disease classification task (ChestX-ray14~\cite{wang2017chestx}) and the pneumothorax segmentation task (SIIM~\cite{siim-acr}) while also evaluating them on zero-shot anatomical understanding tasks, including DNA-test and anatomy correspondence. As shown in Tab. \ref{ablation:components}, each branch contributes uniquely: the extrapolation branch improves fine-tuning, the order correction branch excels in landmark correspondence, and the composition-decomposition branch leads in DNA-test performance. However, simply combining the three learning perspectives by summing the losses results in degraded performance. In contrast, our cyclic training method optimally combines the branches, achieving the best performance on downstream tasks.

\section{Conclusion}

We introduce Lamps, a novel SSL method for visual representation learning that leverages multiple perspectives: extrapolation, order correction, composition and decomposition. These perspectives are integrated with cyclic pretraining to accrue knowledge within a Student-Teacher framework. Lamps has been rigorously tested through comprehensive experiments in various tasks, demonstrating its learned properties in comprehending anatomy and effective transferability, showing significant promise for explainable AI applications in medical image analysis.

\section*{Acknowledgments}
This research has been supported in part by ASU and Mayo Clinic through a Seed Grant and an Innovation Grant, and in part by the NIH under Award Number R01HL128785. The content is solely the responsibility of the authors and does not necessarily represent the official views of the NIH. This work has utilized the GPUs provided in part by the ASU Research Computing and in part by Sol~\cite{jennewein2023sol} and Bridges-2 at Pittsburgh Supercomputing Center through allocation BCS190015 and the Anvil at Purdue University through allocation MED220025 from the Advanced Cyberinfrastructure Coordination Ecosystem: Services \& Support (ACCESS) program, which is supported by National Science Foundation grants \#2138259, \#2138286, \#2138307, \#2137603, and \#2138296. The content of this paper is covered by patents pending.

\bibliographystyle{splncs04}
\bibliography{main}

\end{document}